\documentclass[lettersize,journal]{IEEEtran}

\usepackage{pifont}
\usepackage{color,soul}
\usepackage[ruled]{algorithm2e}
\usepackage{mathptmx} 
\usepackage{booktabs}
\usepackage{multirow}
\usepackage{makecell}
\usepackage[table,xcdraw]{xcolor}
\usepackage{fancyhdr}
\usepackage[normalem]{ulem}
\usepackage[hyphens]{url}
\usepackage[sort,nocompress]{cite}
\usepackage[final]{microtype}
\usepackage[bookmarks=true,breaklinks=true,letterpaper=true,colorlinks,citecolor=blue,linkcolor=blue,urlcolor=blue]{hyperref}

\usepackage{amsmath,amsfonts}
\usepackage{algorithmic}
\usepackage{array}
\usepackage[caption=false,font=normalsize,labelfont=sf,textfont=sf]{subfig}
\usepackage{textcomp}
\usepackage{stfloats}
\usepackage{url}
\usepackage{verbatim}
\usepackage{graphicx}
\def\BibTeX{{\rm B\kern-.05em{\sc i\kern-.025em b}\kern-.08em
    T\kern-.1667em\lower.7ex\hbox{E}\kern-.125emX}}
\usepackage{balance}
\begin{document}
\title{SpikeX: Exploring Accelerator Architecture and Network-Hardware Co-Optimization for Sparse Spiking Neural Networks}

\author{
  Boxun Xu, \textit{Graduate Student Member, IEEE}, Richard Boone, and Peng Li, \textit{Fellow, IEEE} \thanks{Boxun Xu, Richard Boone and Peng Li are with the Department of Electrical and Computer Engineering, University of California at Santa Barbara, Santa Barbara, CA 93106, USA(e-mail: boxunxu@ucsb.edu; richardboone@ucsb.edu; lip@ucsb.edu).}
}



\maketitle

\begin{abstract}
Spiking Neural Networks (SNNs) are promising biologically plausible models of computation which utilize a spiking binary activation function similar to that of biological neurons. SNNs are well positioned to process spatiotemporal data, and are advantageous in ultra-low power and real-time processing. Despite a large body of work on conventional artificial neural network accelerators, much less attention has been given to efficient SNN hardware accelerator design. In particular, SNNs exhibit inherent unstructured spatial and temporal firing sparsity, an opportunity yet to be fully explored for great hardware processing efficiency. In this work, we propose a novel systolic-array SNN accelerator architecture, called SpikeX, to take on the challenges and opportunities stemming from unstructured sparsity while taking into account the unique characteristics of spike-based computation. By developing an efficient dataflow targeting expensive multi-bit weight data movements, SpikeX reduces memory access and increases data sharing and hardware utilization for computations spanning across both time and space,  thereby significantly improving energy efficiency and inference latency. Furthermore, recognizing the importance of SNN network and hardware co-design, we develop a co-optimization methodology facilitating not only hardware-aware SNN training but also hardware accelerator architecture search, allowing joint network weight parameter optimization and accelerator architectural reconfiguration. This end-to-end network/accelerator co-design approach offers a significant reduction of 15.1$\times$$-$150.87$\times$ in energy-delay-product(EDP) without comprising model accuracy. 
\end{abstract}


\section{Introduction}
Spiking neural networks (SNNs), the third generation of neural networks \cite{MAASS19971659}, are models of computation which more closely resemble biological neurons than their conventional non-spiking artificial neutral network (ANN) counterparts. Recent years have witnessed 
great progress in utilizing SNNs for neuromorphic applications \cite{TrueNorth, neuromorphicSNN, DVSG_spiking_SOTA, davies2018loihi} with ultra-low power dissipation \cite{TrueNorth,davies2018loihi} and high accuracy  \cite{TSSLBP,shrestha2018slayer, DBLP:journals/corr/abs-2003-12346}.
Additionally, SNNs are essential tools in the field of computational neuroscience, where mapping and imitating the brain is essential to the exploration of human learning and the advancement of neuroscience \cite{largescalebrainmodel}.   

From a computational perspective, SNNs present two compelling advantages to ANNs due to the nature of their activation function.  First, spiking neural networks inherently operate not only spatially but also temporally due to the temporal nature of the spiking activation function. This temporal affinity gives SNNS the potential to take advantage of highly temporal datasets and problem spaces.  Recent works have shown SNNs can effectively and efficiently learn on neuromorphic datasets such as DVS-Gesture \cite{DVSG, TSSLBP, DVSG_spiking_SOTA}, N-MNIST \cite{NMNIST, TSSLBP}, Neuromorphic-CIFAR10 \cite{CIFAR10_DVS,DBLP:journals/corr/abs-2003-12346}, and others \cite{zhang2015digital, H2MBP, zhang2019spike}.  Additionally, due to the binary nature of spiking activation functions, SNNs show significant potential for high efficiency processing in low power and edge computing applications \cite{WANG202196}.

Despite these significant advantages, far less research effort had gone into acceleration and efficient processing of SNNs than has gone into their ANN counterparts.  While SNNs are similar in general architecture and neuron structure to ANNs, the added temporal dimension and the binary nature of the inputs cause SNNs to run very inefficiently on most or all ANN accelerators, including the more general-purpose accelerators like GPUs and TPUs \cite{5596334, tpupaper}.  
The previous generations of accelerators do not maximally optimize for the unique dataflow opportunities of spiking neural networks, especially as they relate to spatiotemporal sparsity.  

As such, we propose a systolic-array SNN architecture, namely SpikeX, to take into account unique characteristics of spike-based computation. Input and output activations of spiking neurons are binary while weight data are multi-bit. This data disparity favors dataflows that are optimized for reducing expensive weight data movements and maximizing weight sharing. Furthermore, well-trained SNNs exhibit a high degree of spatial and temporal firing sparsity \cite{TSSLBP}, providing a great opportunity for achieving energy and processing efficiency. However, firing sparisty is typically unstructured, i.e., spontaneous firing spikes in an SNN scatter and cluster  with irregular patterns which are not known \emph{a priori}. Exploring such architectural supports that are agile in mapping workloads to the processing element (PE) array and managing the dataflow across time and space. SpikeX is designed with these important considerations in mind while exploring two techniques: \emph{Agile SpatioTemporal Dispatch} and \emph{Activation-induced Weight Tailoring} to deliver significant improvements in energy efficiency and latency.  

Moreover, most SNN training or SNN accelerator design and configurations treat the two as distinct and separate problems despite their interdependencies. Thus, doing so can miss the great benefits resulted from network/accelerator co-optimization.  Recognizing the critical role of firing sparsity in determining the hardware performance of the SpikeX architecture, we develop a set of SNN energy and latency models parameterized in sparsity and network topology, and leverage sparity to bridge between SNN weight parameter optimization and SpikeX reconfiguration. This gives rise to a simple yet effective interface for network/hardware co-optimization, enabling not only hardware-aware SNN training but also hardware accelerator architecture search. Using the combined hardware and software techniques, SpikeX demonstrates up to a 99\% reduction in latency and a 96\% reduction in energy as compared to our baseline.  Additionally, we show a generalizability across a variety of styles of SNNs and the ability to specialize hyperparameters to most optimally match the required computations of a given network. 

\section{Background and Existing Work}
Inspired by research into the behavior of biological neurons, spiking neural networks allow an exploration of biologically inspired computation, as well as a compelling learning framework for neuromorphic, temporal , and low-power applications.  Because SNNs are inspired by and derived from biological neurons, a variety of activation functions have been created over time to most accurately imitate the behavior seen in biology, or to apply a semi-plausible activation function that allows for effective and efficient learning in deep SNNs.  Of the more biologically plausible varieties, the most common is the Hodgkin-Huxley model \cite{Hodgkin1952}, which imitates the behavior of real biological neurons by making mathematical approximations of the ion channels present in somatic and synaptic cell walls.  

However, because of the computational complexity of Hodgkin-Huxley, the simpler discretized Leaky Integrate-and-Fire (LIF) model is most commonly used in state-of-the-art SNN models \cite{TSSLBP, DVSG_spiking_SOTA, shrestha2018slayer}.
As with all spiking neural models, the LIF model operates over a number of timesteps, consisting of two major operations per timestep.  First, the synaptic current at time $t$, $C[t]$ is accumulated from the set of weighted synaptic inputs $x$ as in (\ref{eq:synapticcurrent}) in a similar manner to the weighted linear transformation present in ANNs.  
\begin{equation}
\label{eq:synapticcurrent}
    C[t] = \sum x[t]\times W
\end{equation}
Second, the accumulated synaptic current at time $t$ is put through the spiking activation function, retaining some residual leaky voltage $\lambda u[t-1]$ from time $t-1$ and outputting a spike and resetting if the internal membrane voltage $u[t]$ crosses some threshold voltage $V_{th}$.  Thus, the internal membrane voltage and the spiking output of each neuron are defined by (\ref{eq:membranepotentialupdate}) and (\ref{eq:spikingoutput}), respectively.
\begin{equation}
\label{eq:membranepotentialupdate}
    u[t] = H(V_{th}-(\lambda\times u[t-1] + C[t]))(\lambda\times u[t-1] + C[t])
\end{equation}
\begin{equation}
\label{eq:spikingoutput}
    S[t] = H(V_{th}-(\lambda\times u[t-1] + C[t]))
\end{equation}
Because the Heaviside step function $H(x)$ is not differentiable, training methodologies and derivative approximations vary between different leading methodologies.  Despite this, SNNs have demonstrated impressive performance in a variety of fields including computer vision \cite{TSSLBP, shrestha2018slayer, DVSG_spiking_SOTA} and speech recognition \cite{H2MBP, zhang2015digital, zhang2019spike}.

The above standards lead to a number of important properties among spiking neural networks.  Most importantly, the input and output of each layer of the network, the spike $S$ is binary at each time point.  This means that instead of requiring a full multiply and accumulate for each input to the neuron, the summation can be done by a simple conditional accumulate.  
The uniquely binary inputs and outputs of an SNN lead to a different set of tradeoffs for data movement in accelerators.  In ANNs, as a general rule, activation movement is more costly in convolution layers, and weight movement is more costly in fully connected layers.  

Because of SNNs' unique characteristics and temporal nature, the optimal dataflow for an SNN may significantly differ from that of an ANN with the same overall structure.  Additionally, most SNNs exhibit a high degree of spatial and temporal sparsity.  In simple terms, at a given timestep, most neurons are not firing, and for a given neuron, only a small number of timesteps exhibit firing activity.  This means that for most SNNs, the true degree of sparsity is much higher than that of ANNs, even prior to the use of any methods for lowering the computational complexity of inference in the network.  We demonstrate this in Figure~\ref{fig:dvsg_before_after}a which shows the number of spikes of the average neuron in a network trained using state-of-the-art SNN training techniques.  As noted in \cite{Foldiak:2008} biological neurons see similar sparsity levels relative to their maximum firing rate.

\begin{figure*}[th]
    \centering
    \includegraphics[width=1\textwidth]{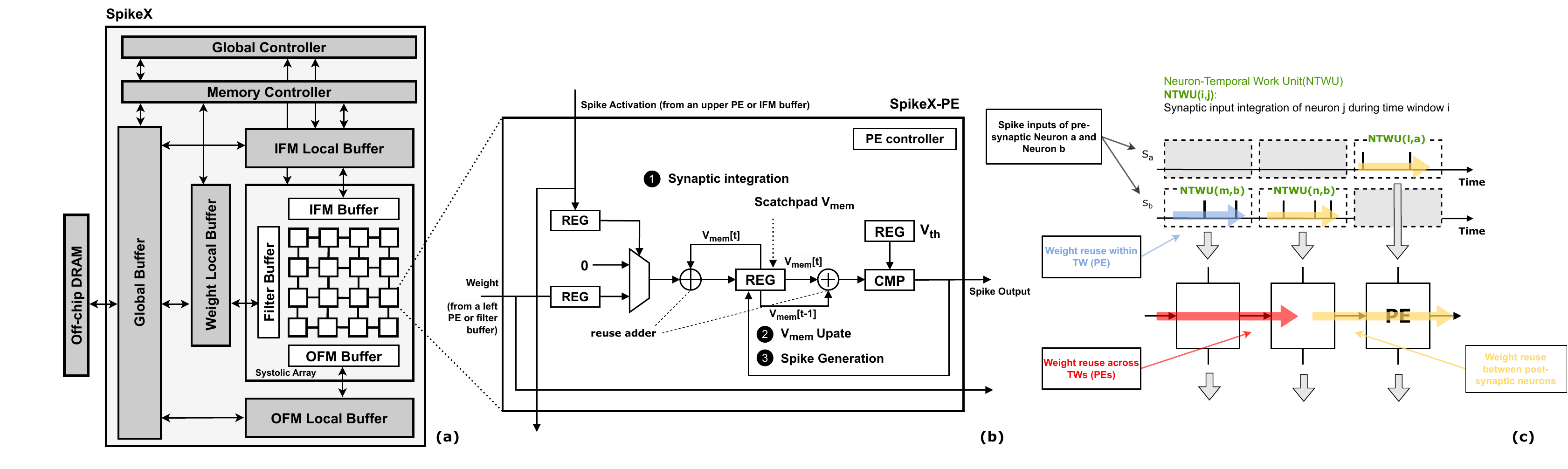}
    \caption{(a) The proposed SpikeX hardware Architecture (b) The schematic design of Processing Element (PE) in the systolic array (c) A comprehensive weight sharing scheme that operates at three levels: (1) within the same time window (TW), (2) across multiple TWs, and (3) across post-synaptic neurons}
    \label{fig:Hardware}
\end{figure*}

\subsection{Existing SNN Hardware Accelerators}
While a large amount of research has gone into fast and efficient hardware accelerators for ANNs \cite{guo2018survey}, these accelerators are not efficient for SNNs because the don't take advantage of the previously discussed unique characteristics of SNNs.  Of the existing SNN accelerator literature, a few must be discussed and evaluated for their strengths and weaknesses.  

IBM's TrueNorth chip \cite{TrueNorth} is one of the first large industrial full-chip designs for a neuromorphic hardware accelerator of SNNs.  It is capable of handling up to 1 million digital neurons with 256 million synapses with a high degree of efficiency.  In a similar manner, Intel's Loihi chip \cite{davies2018loihi}, and associated expansion systems \cite{loihi2}, provide a neuromorphic generic hardware accelerator for SNN computation, which demonstrates significant power and latency gains as compared to running on conventional ANN accelerators.  While such accelerators demonstrate very effective benchmarks, the purely neuromorphic architecture falls short on large or dense networks, where weights cannot always be fully loaded onto the hardware.  

Aside from Truenorth and Loihi, a number of hardware SNN accelerators have been proposed.  \cite{wang2020sies} Uses an FPGA architecture and an ANN to SNN conversion to efficiently process convolutional layers in SNNs.  \cite{painkras2013spinnaker, mayr2019spinnaker, furber2014spinnaker}  Use a number of interacting ARM cores with specialized routing for large-scale SNN simulations, specifically targeting imitation of a human brain.  \cite{Spinalflow} uses a combination of compression and time-parallel processing to efficiently process each neuron on a parallel array of processing elements.  Of these, \cite{wang2020sies, SpiNNaker:IEEE:2014, Spinalflow} do not demonstrate dataflows for parallel time processing and  \cite{Spinalflow} is only applicable to the high restricted family of SNNs in which each neuron only fires once.  Recent work on systolic array SNN accelerators \cite{PTB, Lee_ICCD_2020} have shown that SNN optimal dataflows are not necessarily the same as those of ANNs, and often include parallel processing across the time dimension, methods which are not demonstrated in the existing hardware. \textcolor{black}{
\cite{liang2021h2learn} designs look-up table (LUT)-based PEs to handle efficient backpropagation-through-time (BPTT) training accelerator; Furthermore, \cite{yin2022sata} proposes a systolic-array-based sparsity-aware training accelerator to handle the three groups of sparsity (spike S, the gradient of firing function, and the gradient of membrane potential) when back-propagation.
\cite{yin2024mint} proposes a multiplication-free quantized LIF model, sharing the quantization scaling factor between weights and membrane potentials, eliminating the need for multipliers required in conventional uniform quantization. \cite{xu2024trimming} proposes a heterogenous mixed-precision compression framework for large-scale spiking vision transformers.
\cite{bhattacharjee2024snns} identifies and addresses key roadblocks to efficient SNN deployment on hardware.\cite{xu2024spiking} and \cite{xu2024towards} explore efficient deployments on 3D-integrated hardware for self-attention mechanisms and mixture-of-expert schemes. \cite{yin2024workload} proposes a workload-balanced pruning to alleviate the weight imbalance issue when weight has already had a high sparsity.}

\subsection{Challenges and Opportunities in SNN \\  Hardware Acceleration}
\textbf{Activation/weight data disparity and temporal parallel processing.} Unlike in the conventional artificial neural networks (ANNs) where input/output activation and weight data are both multi-bits, activations  are one bit (binary) while the bit-width may vary from 8 to 32 bits in an SNN \cite{Lien_2022}. Hence, it is desirable to develop SNN dataflows specifically minimizing  expensive weight data movement. 
While recognizing this data disparity,  \cite{PTB} explores temporal processing in which synaptic input integrations over multiple time points within a time window share the same weights, giving the rise to the idea of parallel time batching (PTB). While PTB goes beyond the time-serial processing of most of other SNN accelerators, it does not fully  tap into the opportunities of weight sharing and weight data movement minimization. In addition, \cite{PTB,Lee_ICCD_2020} suffer from poor PE utilization when temporal sparsity is high, which is typically the case for well-trained SNNs.  

\textbf{Unstructured firing sparsity.}
As a key characteristic of spike-based computation, SNNs often exhibit a high degree of firing sparsity in both space and time. 
For instance, only 0.0001\% of the CONV3 layer neurons  of a well-trained SNN model for neuromorphic DVS-Gesture dataset \cite{amir2017low} fire $>$ 150 spikes over 300 time points. Sparsity  offers a great opportunity for minimizing expensive weight data access and avoiding redundant PE processing on the hardware accelerator.  However, sparse firing patterns in an SNN are highly irregular: not all neurons fire at a given time (spatial irregularity), and the firing count and firing times of each neuron can vary dramatically (temporal irregularity).   Spatial and temporal irregularities make it challenging to explore sparsity to improve energy efficiency, latency, and PE utilization of the accelerator. 

\textbf{Proposed Work.} The proposed SpikeX systolic-array architectures goes beyond  the PTB architecture  \cite{PTB} significantly. It  explores three-levels of weight sharing not only in time but also in space via the proposed \emph{Agile SpatioTemporal Dispatch} approach. In addition,  SpikeX utilizes \emph{Activation-induced Weight Tailoring} to minimize weight data access by avoiding loading weights associated with zero-valued input activations across the memory hierarchy. Finally, this work enables network/hardware co-optimization via hardware-aware training and accelerator architecture search, leading to additional large energy-delay-product improvements without compromising model accuracy.

\section{SpikeX Architecture}\label{sec:HA}

\subsection{Overview of the SpikeX Architecture}
Our SpikeX architecture consists of a systolic array with hierarchical memories and buffers for data storage as shown in Figure~\ref{fig:Hardware}(a). SpikeX accelerates a deep SNN in a layer by layer manner. As in Figure~\ref{fig:Hardware}(b), each Processing Element (PE) consists of one reusable accumulator, one comparator, a small scratchpad memory, registers, and PE control logic. We adopt a three-level memory hierarchy consisting of (1) off-chip RAM, (2) a global buffer(GLB) (3) a double-buffered L1 buffer(LBUF)\cite{double-buffer, PE}. 

To process the postsynaptic neurons in a given layer, input activation and weight data are fed from top to bottom, and from left to right, respectively to the PE array, leveraging the high-data bandwidth along the edges of the systolic array.  During each processing cycle, a PE processes a postsynaptic neuron  over multiple time points within a time window (TW) following the integrate-and-fire model (\ref{eq:synapticcurrent}, \ref{eq:membranepotentialupdate}, \ref{eq:spikingoutput}) in three steps.  In step \ding{182}, the activations  across various input channels and time points are integrated based on the corresponding weights,  passed from the left PE from  or the filter buffers. Upon the completion of input integration  at all time points within the TW in step \ding{182}, $V_{mem}[t]$ at each time point $t$ is updated by sequentially considering the value of  $V_{mem}[t-1]$ in step \ding{183}. In step \ding{184},  a spike output is generated if the updated $V_{mem}[t]$ is above the firing threshold voltage $V_{th}$ for each time point $t$.  

The time window size (TWS) specifies the number of timesteps a single PE is expected to process for a given postsynaptic neuron, and is reconfigurable. This lets SpikeX to be adaptable to different network and layer configurations, allowing the network/accelerator co-optimization discussed in Section~\ref{sec:HT}. 

\subsection{Agile SpatioTemporal Dispatch}

Spike-based computation must operate across both spatial and temporal dimensions.  For this reason, we have two separate goals for loading data onto the accelerator. First, these spatio-temporal workloads should be packed with a proper granularity to be mapped onto different PEs. Second, the workloads should be flexibly mapped to the PE array to maximize multi-bit weight data reuse and PE array utilization, and to exploit unstructured firing sparsity. The proposed Agile SpatioTemporal Dispatch approach achieves these two goals respectively by packing the workloads into multiple neuro-temporal work units (NTWUs), and mapping the packed NTWUs on the array in a manner to maximize data reuse and array utilization.   

\begin{figure}[htbp]
    \centering
    \includegraphics[width=0.5\textwidth]{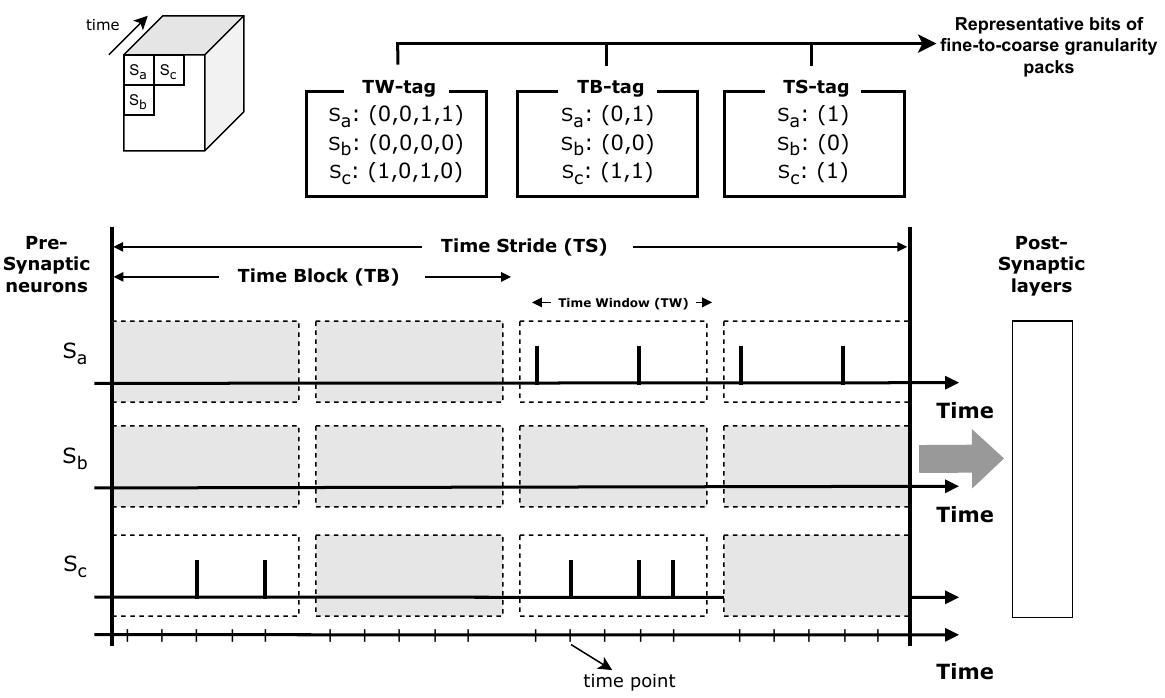}
    \caption{Four temporal granularities: time point, time window (TW), time block (TB),  and time stride(TS), and the corresponding activity tags.}
    \label{fig:tag}
\end{figure}

In order to define an NTWU, and thus to evaluate computational efficiency, we first define four levels of temporal granularity.  Figure~\ref{fig:tag} shows these four temporal granularities with a single  time point  being the finest granularity: \\
\textbf{Time window (TW):} a TW consists of TWS time points and constitutes the basic temporal granularity of NTWUs for workload mapping to the PE array.  TWS remains consistent across all PEs during computation.  \\
\textbf{Time block (TB):} A TB consists of multiple time windows, and  is sized relative to the corresponding cache memory size at  different levels of the memory hierarchy. \\
\textbf{Time stride (TS):} TS represents the entire time horizon over which the SNN inference runs;  TS is equally split into multiple time blocks (TBs).
In Figure~\ref{fig:tag}, the TS is broken into four TBs, each containing two TWs.  

\subsubsection{Neuro-temporal Work Units (NTWU)}
$NTWU(n, t_w)$ defines the basic spatiotemporal workload unit for processing a postsynaptic neuron $n$ over a specific time window $t_w$.  $NTWU(n, t_w)$ corresponds to the amount of computational work required to process the neural activities and produce the spike outputs for neuron $n$ over all time points in time window $t_w$ based on the integrate-and-fire model (\ref{eq:synapticcurrent}, \ref{eq:membranepotentialupdate}, \ref{eq:spikingoutput}). Here, the spatial granularity of NTWUs is a postsynaptic neuron while the temporal granularity is a time window.

For each temporal unit, we define an activity tag which is zero if and only if there is no presynaptic input firing activity within the unit. In addition, we perform hierarchical tagging in which the bitwise or of the lower-level activity tags forms the tag one level up.   
At different levels of granularity, data or work with a zero-valued activity tag are not loaded or processed. Such tag-based zero-skipping enables exploitation of spatiotemporal sparsity in SNNs.  
Among these temporal granularities, TB defines the temporal granularity for data storage/loading, which is described in Section~\ref{sec:tailoring}.

Using time window as the granularity of NTWUs and packing multiple time points into a time window has several advantages \cite{PTB}. 
SNNs tend to demonstrate a very high degree of firing sparsity as shown in previous studies \cite{TSSLBP, DL_SNN_2019}. Irregularity of firing patterns, however, renders exploitation of sparsity for weight data reuse and zero-skipping redundant computation challenging.  Packing multiple time points into a time window addresses these two issues. First of all, firing spikes have a strong tendency to cluster as opposed to spread out uniformly in time.  This is reflected in Figure~\ref{fig:dvsg_before_after}, which shows a high degree of sparsity not only at the individual spike level but also in terms of packed time windows, i.e.,  there are only a very few active time windows in which firing spikes cluster. This implies that zero-skipping can be more structurally performed at NTWU level by processing only the NTWUs with a non-zero activity tag.  
Furthermore, as an NTWU is mapped to a PE, the same weight data can be shared to perform synaptic input integration, the most computationally intensive step of the LIF model,  across all time points in the time window. This avoids expensive repeated access to the same weights as would be the case in the conventional time-serial processing where synaptic input integration takes place time step by time step.


\subsection{Agile NTWU Dispatch with Three-Level of Weight Reuse}
The weight data sharing within the time window of a NTWU constitutes the first level of weight data reuse in SpikeX. We explore two higher-level weight data sharing, as illustrated in Figure~\ref{fig:Hardware}, by agile dispatch of NTWUs.  

\begin{figure*}[ht]
    \centering
    \includegraphics[width=\textwidth]{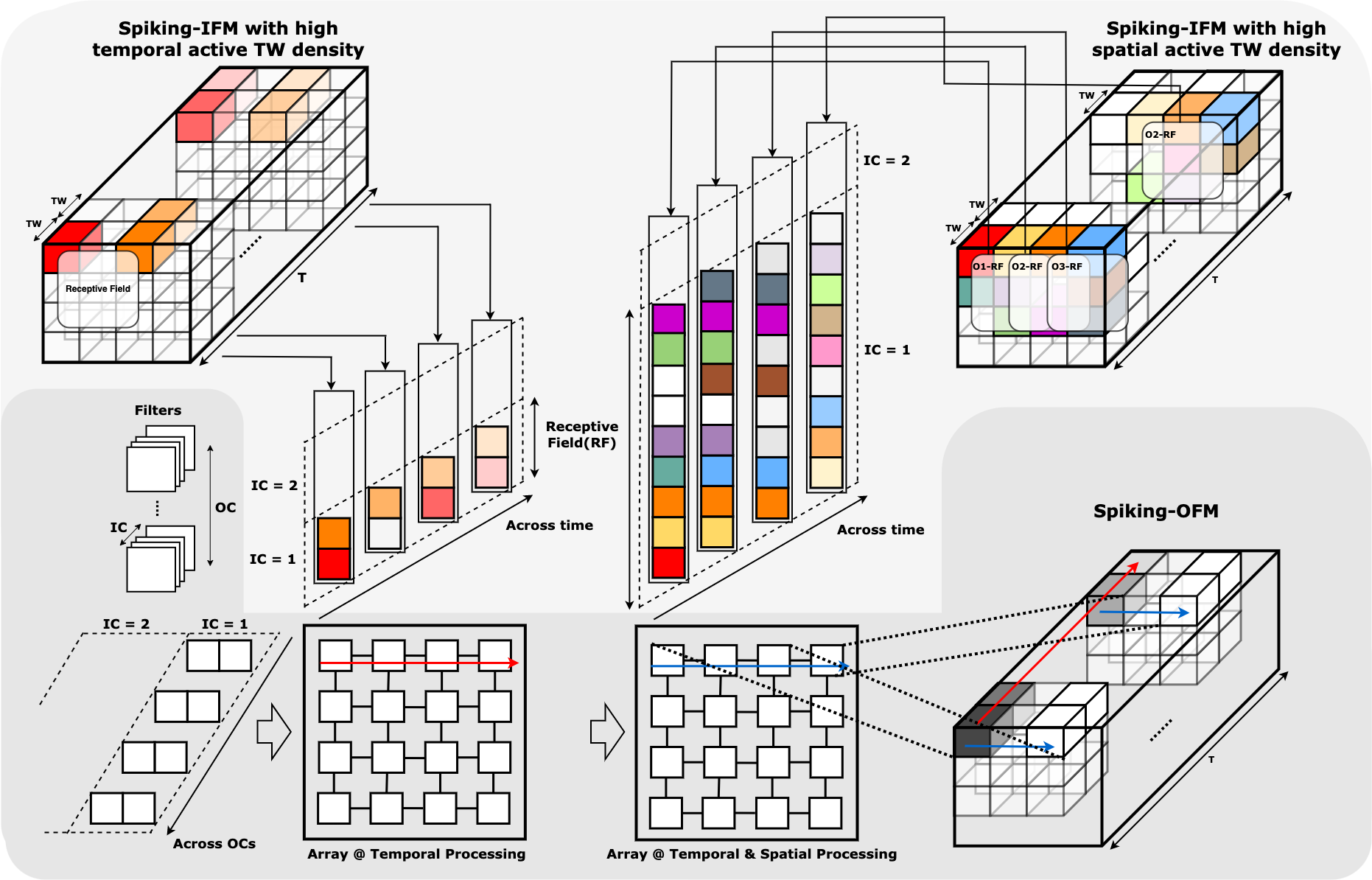}
    \caption{Agile SpatioTemporal Dispatch. (Left) High temporal density mode, where NTWUs across TW grids are dispatched onto distinct PE columns. (Right) High spatial density mode, where NTWUs across various spatial grids and TW grids are dispatched onto different PE columns.}
    \label{fig:dispatch}
\end{figure*}

Among a pool of active NTWUs, the dispatch scheduler can operate in two modes: \emph{high temporal NTWU density} and \emph{high spatial NTWU density}. 

The first (default) mode is activated when the average number of active NTWUs per neuron is more than the array width. 
The scheduler maps a pool of NTWUs such that ones for  the same postsynaptic neurons are assigned to the same row of the array while different rows process NTWUs of different postsynaptic neurons. In this case, 
input activities of different time windows are fed into different columns (PEs) of a row from the top of the array. In the meantime,  the same weight data is fed into the row from the left and shared among all PEs in that row. This constitutes the second level of weight data reuse. The high temporal NTWU density ensures that nearly all columns of the PE array are occupied, leading to high utilization, as shown in the left of Figure~\ref{fig:dispatch}.

In addition to the first two levels of weight data sharing enabled in the high temporal NTWU density mode, in the \emph{high spatial NTWU density} mode (right of Figure~\ref{fig:dispatch}), the schedule further tries to occupy all columns of the array by mapping active NTWUs of different neurons into the same row as long as such NTWUs require the same weights. This mode is particularly effective for convolutional layers, where the same filter data is used to process different postsynaptic neurons in the layer. This allows for the third level of weight data reuse between PEs processing different post-synaptic neurons. Note that convolutional layers are often the dominant contributors to the latency and energy dissipation of the accelerator.

\subsection{Activation-induced Weight Tailoring} \label{sec:tailoring}

While agile spatiotemporal dispatch improves PE utilization and weight reuse  across different TWs and post-synaptic neurons, the dataflow can be further optimized to minimize weight data access by avoiding loading redundant weight data. We achieve this goal by the proposed \emph{activation-induced weight tailoring}. 

Activation-induced weight tailoring is applied at each level of the memory hierarchy to \emph{tailor} the weights that correspond to zero-valued input activation tags. The spatiotemproal binary input activation and spatial (static) multi-bit weight data are stored separately in each cache. The input activations are stored  in multiple \emph{spatiotemproal memory blocks} (SP-MB). Each SP-MB encapsulates input data for a number of postsynaptic neurons over multiple time blocks (TB). The size of a SP-MB is properly chosen relative to the total cache size at that level and limited by the cache size one level below. We associate an activity tag with each SP-MB, and it is simply the bitwise OR of the tags of all TBs in the SP-MB. Hence, the activity tag is zero if and only if there is no active input activation in the entire SP-MB. While loading the data to the lower-level cache, the memory controller only fetches the data associated with active SP-MBs with a non-zero activity tag. 

\begin{figure}[ht]
    \centering
\includegraphics[width=0.5\textwidth, clip,trim={9.3cm, 6cm, 1.5cm, 1cm}]{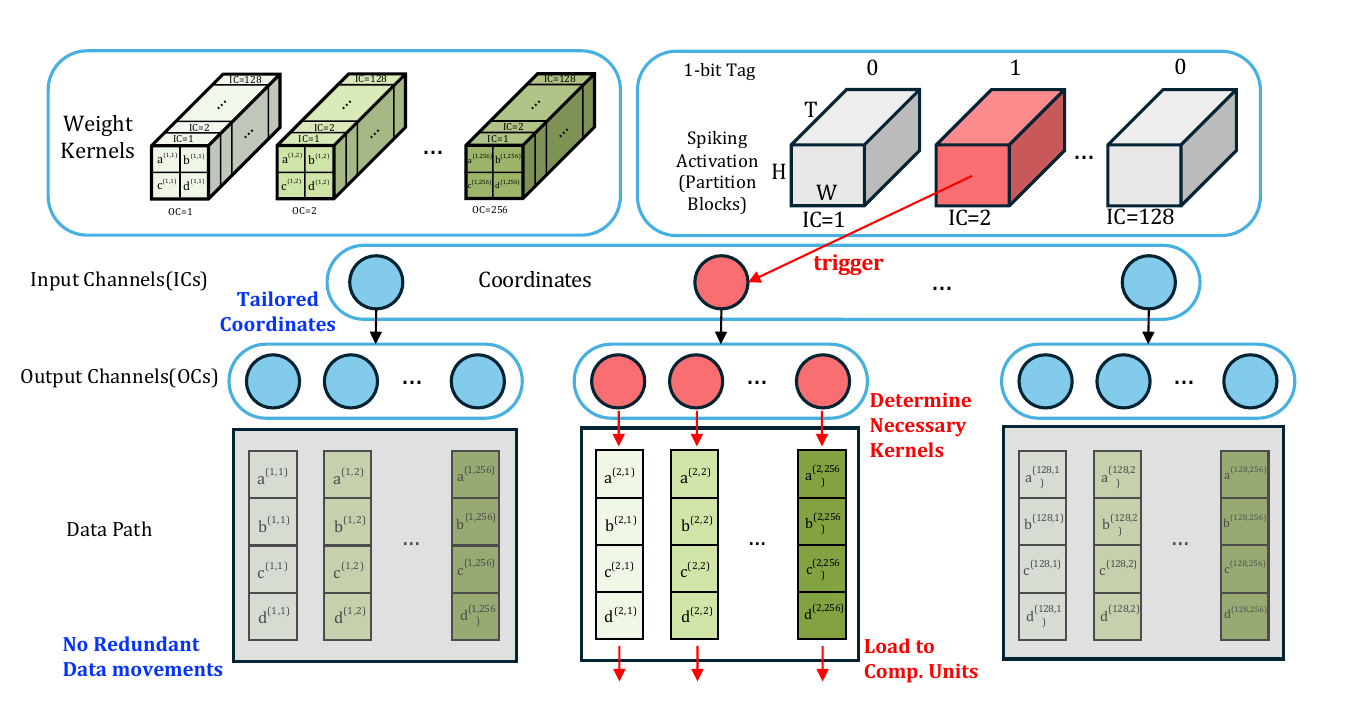}
    \caption{Activation-induced weight tailoring during determining necessary weight kernels loaded to computation units.}
    \label{fig:tailor}
\end{figure}

In practice, input activation data can be partitioned across different SP-MBs in a variety of ways at a given memory levels and for different types of layers, e.g., densely-connected vs. convolutional layers. Figure~\ref{fig:tailor} illustrates the SP-MBs and their fetching to the lower-level cache for convolutional layers. Here, each SP-MB encapsulates input activations from  a single input channel (IC) to a set of post-synaptic neurons over multiple time blocks. In this case, only the weight data corresponding to the input channel of an active SP-MB are loaded to the lower-level cache while others are tailored.   
This approach can effectively reduce energy consumption and latency by aggressively tailoring weight access, especially when the firing activities of the network are highly sparse. 


\section{Network and Accelerator Co-Optimization for Spiking Neural Networks(SNNs)}\label{sec:HT}

In Section~\ref{sec:HA}, we present a hardware architecture and control framework that can efficiently process a sparse spiking neural network. Our primary objective is to optimize both the network and the architecture to achieve efficient processing with minimal loss in accuracy. To this end, we propose two significant approaches. Firstly, we demonstrate that the network can be optimized for efficient processing on the fixed hardware architecture via \emph{hardware-aware training} (SpikeX-HT) while preserving its accuracy. SpikeX-HT gives rise
to a simple yet effective method for training SNNs while considering the underlying accelerator platform.  
Taking one step further, we show that we can simultaneously optimize the SNN weight parameters and the reconfigurable parameters of the SpikeX accelerator via \emph{hardware architecture search} (SpikeX-HAS) for efficient network processing.

\begin{figure*}[ht]
    \centering
    \includegraphics[width=\textwidth, clip, trim={5cm 5cm 7.5cm 1cm}]{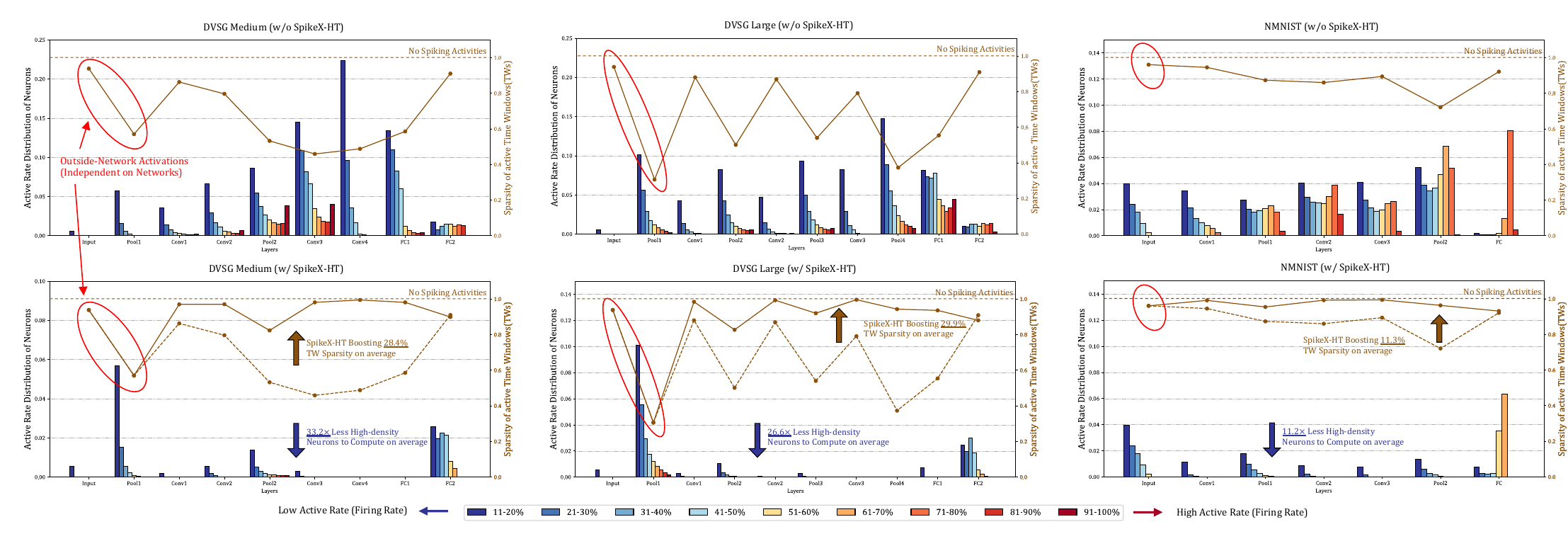}
    \caption{An analysis of realistic layer-wise firing activities of LIF neurons and structured sparsity of packed spiking time windows(TWs) with and without SpikeX-HT, on DVSG medium, DVSG large and NMNIST.}
    \label{fig:dvsg_before_after}
\end{figure*}

\subsection{Hardware-Aware Training (SpikeX-HT)}

Our main objective, when working with a fixed hardware architecture, is to train the network to be processed efficiently while maximizing accuracy. 


\begin{figure}[h]
    \centering
    \includegraphics[width=0.45\textwidth, clip, trim={2cm 4cm 3cm 3.5cm}]{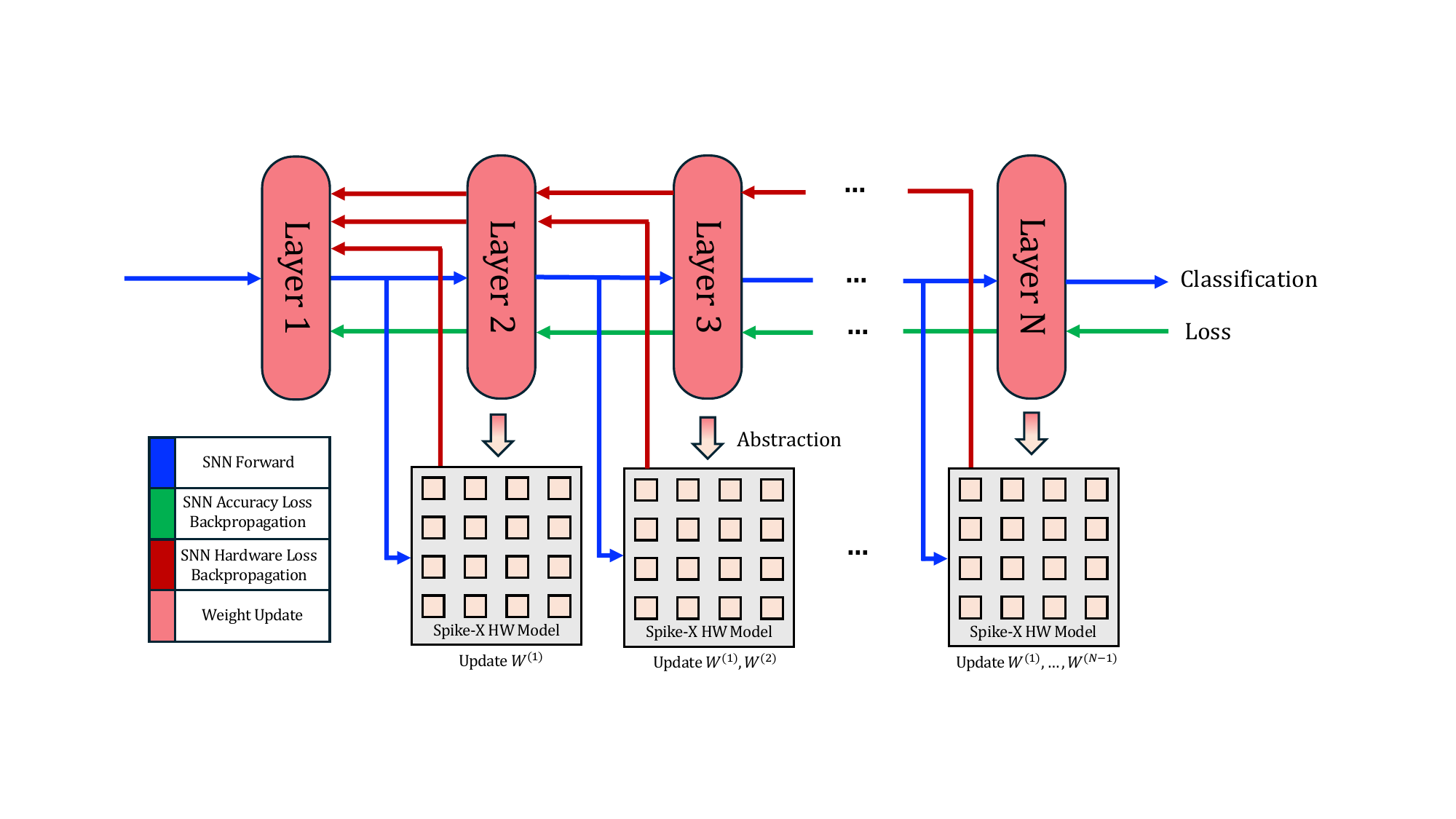}   \caption{The backpropagation of hardware-aware training via sparsity.}
    \label{fig:BP}
\end{figure}

To this end, we introduce two significant learning terms: $L_{acc}$, which represents the accuracy loss of the network when classifying a known dataset, and $L_{HW}$, which captures the combined Energy Delay Product (EDP) of each layer, given the structure of the sparse spikes on the layer input. Our ultimate optimization goal is to minimize the total loss of the network, as in equation (\ref{eq:loss_tot}), where $\beta$ is a parameter that determines the contribution of $L_{HW}$ to the overall loss. This parameter allows us to tailor the optimization to prioritize either more sparse and efficient computation or higher accuracy.  Furthermore, the primary modifiable parameters of the network are the weights, and because the weights are deterministic of each loss function, we can treat them each as functions of the weights $L_{tot}(W) = L_{acc}(W) + L_{HW}(W)$.  Thus our loss $L_{tot}$ can be tuned to optimize for both network accuracy and hardware efficiency while working effectively alongside traditional backpropagation-based learning methods.  
\begin{equation}
\label{eq:loss_tot}
    L_{tot} = L_{acc} + \beta L_{HW}
\end{equation}


Given the complexity of the hardware, even with simulation, exact energy-delay-product (EDP) measurements for a layer given some input take significant time and are far too complex to run simultaneously with network training.  To solve this, an approximate metric must be used to best fit some $L_{HW}$ to EDP.
As demonstrated in Figure~\ref{fig:Sparsity_EL} and in \cite{PTB}, the sparsity of the network plays a critical role in determining the efficiency, i.e., EDP, of the accelerator. We denote a measure of firing sparsity of the network by $S_p$.
Because SpikeX evaluates and skips activations and weights on the level of a time window, we treat the time window sparsity, that is, the number of time windows that contain at least one spike as our primary variable from which to determine approximate EDP. Recognizing the fact that $S_p$ is a function of the SNN weight parameters $W$, i.e.,  $S_p = S_p(W)$, we make use of $S_p$ as a proxy of the hardware accelerator for simplified hardware-aware training. 
For this purpose, we fit a piecewise linear model to the approximate EDP per filled time window based on well-evaluated simulation data.
Such a model allows us to formulate our hardware loss as a function of $S_p$, which is then a function of weights.  Thus, we evaluate $L_{HW}$ as $L_{HW}(W) = EDP(S_p(W))$ where $S_p(W)$ represents the spiking activity of a layer given some weights $W$.  

\begin{figure*}[ht]
    \centering
    \includegraphics[width=\textwidth, clip, trim = {1cm 7cm 1cm 2cm}]{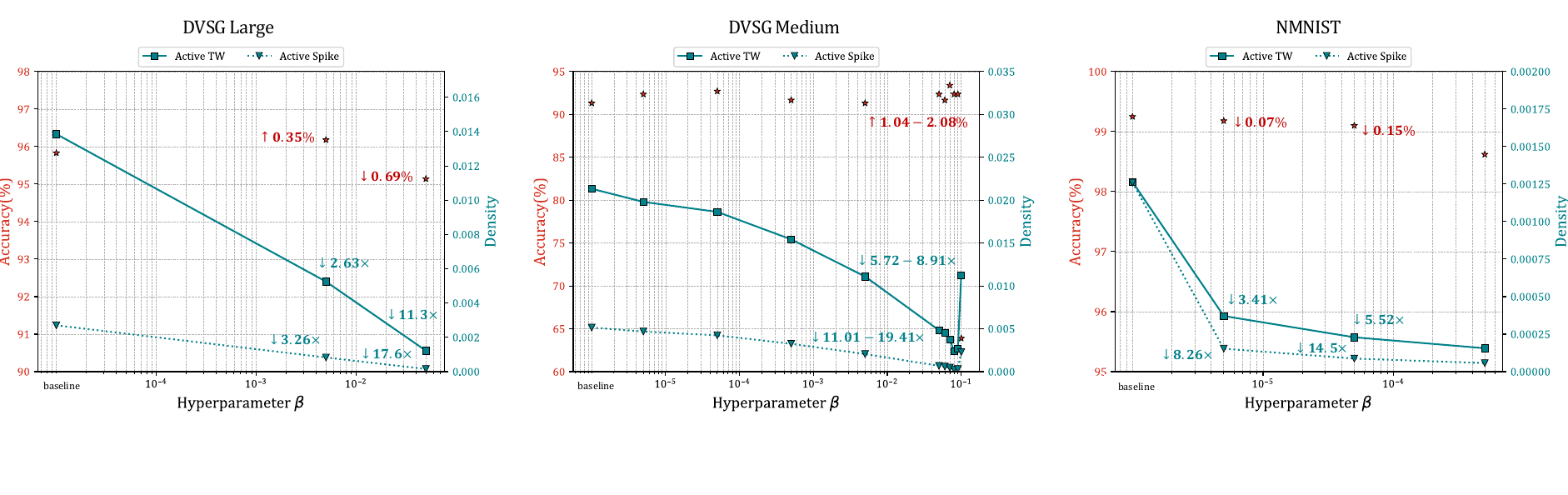}
    \caption{Accuracy, active spike density and active time window(TW) density across different hyperparameter $\beta$ for DVS Gesture Large, DVS Gesture Medium\cite{DVSG} and NMNIST\cite{NMNIST} on the architectures as illustrated in Table~\ref{table:config}.}
    \label{fig:sf_accuracy}
\end{figure*}

With this loss function fully formulated, we can now train the network using backpropagation, with the hardware loss inserted at the output of each layer as can be seen in Figure~\ref{fig:BP}.  For the spiking activation function derivation, we use \cite{TSSLBP} with the temporal cutoff from \cite{boone2021} applied.  Training for the network takes two significant steps. First, the network is pretrained without using the hardware loss and using a warm-up function to induce widespread activation as in \cite{TSSLBP, DVSG_spiking_SOTA} and others. After pre-training, the warm-up function is removed, and the network is trained with HW loss included.  In Figure~\ref{fig:sf_accuracy} we show our training accuracy across a sweep of our sparsity parameter $\beta$ from $5e-05$ to $1e-01$ for three architectures on two datasets. Combined with Figure~\ref{fig:dvsg_before_after} we show a significant decrease in sparsity with little to no effect on accuracy. In fact, in some cases, moderate training levels can increase accuracy whilst simultaneously decreasing sparsity. Finally, we show layerwise sparsity reduction in Figure~\ref{fig:dvsg_before_after} where we can see that after training, almost all neurons fire less than 10\% of the time.

\subsection{Hardware Architecture Search \\ (SpikeX-HAS) }
While we training for a  fixed architecture can demonstrate significant reductions in EDP, the reconfigurable nature of the accelerator means that the hyperparameters of the hardware accelerator can also be jointly optimized with the SNN network weights $W$, leading to the more comprehensive network/hardware co-optimization approach SpikeX-HAS. 

\begin{figure*}[ht]
    \centering
    \includegraphics[width=1\textwidth]{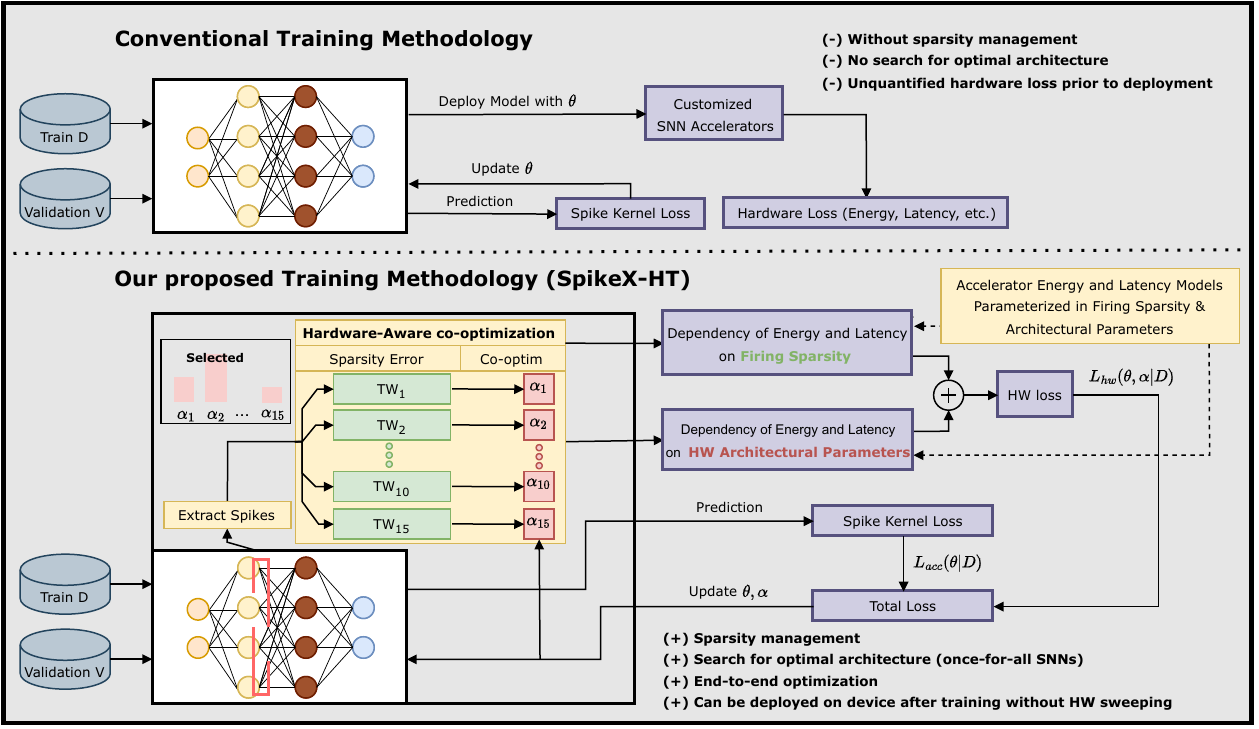}
    \caption{(a) The conventional SNN training methodology (b) Our proposed training methodology Spike-HAS which involves hardware-aware training and hardware architecture/network joint optimization.}
    \label{fig:Search}
\end{figure*}

We adapt an algorithm originally generated for neural architecture search \cite{NAS_survey}. The key distinction here is that architecture search is with respect to the reconfigurable parameters of the accelerator as opposed to that of the neural network model.  As shown in Figure~\ref{fig:Search}, SpikeX-HAS is based on solving the following two-level optimization problem:
\begin{eqnarray}\label{eq:bi_opt}
	&& \quad \quad \quad min_{ \alpha} L_{tot}( \alpha, W^*(\alpha),  S_p(\alpha, W^*( \alpha) )) \\
	&&  s.t.  \quad W^*(\alpha) =  arg_{W}min{L}_{tot}( \alpha, W,  S_p( \alpha,  W(\alpha) ))  
\end{eqnarray}
Note that $\alpha$ corresponds to  discrete accelerator architectural configuration parameters, e.g., time window sizes for different layers, which can be generalized to include other accelerator parameters such as  bit resolutions. We employ continuous-valued parameter relaxation for efficient gradient based solution to (\ref{eq:bi_opt}). 
We use continuous-valued $\hat{\alpha}^v$  as the categorical choice of architectural configuration parameter $v$, e.g., a particular TWS. We construct a hypernet  of the SNN hardware loss function $L_{HW}$ as the sum of  all mappings of the SNNs onto the accelerator under different reconfigurations weighted by the respective the continuous-valued categorical choices (selection probabilities) represented by $\alpha$. The hypernet is optimized by gradient-based optimization, and the final architecture is constructed by choosing the choices with the highest selection probabilities at the end of training. 
With this, we also parameterize the hardware loss by TW size as $L_{HW}(W, \alpha) = EDP(S_p(W, \alpha))$. 

\begin{table*}[h]
\centering
\caption{Comparison of existing and our SNN accelerators and optimization methodology.}
\begin{tabular}{c|c|c|c|c|c}
\hline
& \thead{\textbf{Applicability}} & \thead{\textbf{Parallel Processing}} & \thead{\textbf{Sparsity Handling}} & \thead{\textbf{HW-aware Training}} & \thead{\textbf{Arch Optimization}} \\
\hline
TrueNorth\cite{TrueNorth}, Loihi\cite{davies2018loihi} & High & No & Yes & No & No \\
\hline
Spinalflow\cite{Spinalflow} & Low  & Spatial & Yes & No & No \\
\hline
PTB\cite{PTB} & High & Temporal & Yes & No & Limited \\
\hline
SpikeX(Ours) & High & \makecell[c]{Agile Spatial \\ and Temporal} & \makecell[c]{Coarse-to-grain \\ granularity} & Yes & Yes \\
\hline
\end{tabular}

\label{table:snn_accelerator}
\end{table*} 
We compare the proposed SpikeX architecture and optimization methodology with other SNN acceleartors in Table~\ref{table:snn_accelerator} where we simply use $\alpha$ to represent the choice of Time Window Size (TWS), thus optimizing TWS simultaneously with the standard network-level training.

\section{Evaluation Methodology} 
We introduce an analytic architecture-level simulator, adapting techniques similar to \cite{PTB, Lee_ICCD_2020} to evaluate memory access, energy consumption, and latency.  The simulator enables the tracing of data movement and usage for evaluating both energy dissipation and latency. 

\subsection{Modeling of systolic array and Hierarchical Memory}
 We use a fixed  8-by-8  systolic array as the default setting, which is similar to that in \cite{Eyeriss, Spinalflow}. We adopt a three-level memory hierarchy for memory-intensive computation in neural networks \cite{PTB, Eyeriss, double-buffer, fan2024aimmi}. Each level of memory is double-buffered\cite{double-buffer, Scalesim} to hide latency and is partitioned to store different types of data. The size of the global buffer and local buffer is set as 54KB and 2KB, respectively, and the DRAM bandwidth is set to 30GB/sec. The spiking IFmap/Filters/OFmaps/Psums data are transferred in blocks at different granularities between distinct memory units. We analyze the impact of agile spatiotemporal dispatch, accelerator reconfiguration parameters (hyperparameters), and the proposed optimization techniques. 

\subsection{Performance Modeling}
We introduce an approach to model the performance of memory access, latency, and energy dissipation. The simulator generates a cycle-level data trace of activation, filters, and outputs for each level of memory. To evaluate memory access, the simulator generates mapping orders through data scheduling for a given network configuration. With pre-determined data loading sequences, memory access can be evaluated accordingly. For instance, when loading 3-by-3 8-bit weight data from the global buffer to the local buffer, the counts for read operation of the global buffer and the write operation of the local buffer are incremented by 9 8-bits data access. To evaluate latency, we account for the worst-case scenario between data access and array computation, as the latency of memory access and the latency of array computation are hidden by the pipeline, and the array computation can operate almost stall-free\cite{PTB}. The total latency is calculated by summing the latencies of all operations. To evaluate energy dissipation, we use traces of read/write operations at each level of memory and the total number of arithmetic operations of the PEs. We employ a standard modeling strategy and configure CACTI\cite{CACTI} for 32nm CMOS technology to evaluate energy dissipation of memory. We do this by multiplying the number of accesses based on the read/write traces by the energy per memory access at each level of memory. The arithmetic energy consumption is evaluated by multiplying the total number of accumulation operations for a specific network by the energy per accumulation operation\cite{PE}.

\begin{table}
\centering
\caption{Network Configurations of spiking neural networks used as benchmarks.}
\label{table:config}
\begin{tabular}{ccccccc}
\toprule
\rowcolor[HTML]{EFEFEF} 
\shortstack{\textbf{Dataset}} & \textbf{Layer} & \textbf{H} & \textbf{R} & \textbf{E} & \textbf{C} & \textbf{M} \\ \midrule
\multirow{6}{*}{\shortstack{DVS-Gesture Medium\\timestep: 300}} & CONV1 & 64 & 32 & 5 & 2 & 8 \\
& CONV2 & 32 & 8 & 5 & 8 & 16 \\
& CONV3 & 8 & 8 & 5 & 16 & 16 \\
& CONV4 & 8 & 10 & 5 & 16 & 16 \\
& FC1 & 1 & 1 & 1 & 1600 & 512 \\
& FC2 & 1 & 1 & 1 & 512 & 11 \\\midrule
\multirow{6}{*}{\shortstack{DVS-Gesture Large\\ timestep: 300}} & CONV1 & 32 & 16 & 3 & 2 & 32 \\
& CONV2 & 16 & 8 & 3 & 32 & 64 \\
& CONV3 & 8 & 4 & 3 & 64 & 128 \\
& FC1 & 1 & 1 & 1 & 2048 & 256 \\
& FC2 & 1 & 1 & 1 & 256 & 11 \\\midrule
\multirow{4}{*}{\shortstack{N-MNIST\\ timestep: 30}} & CONV1 & 34 & 16 & 3 & 2 & 12 \\
& CONV2 & 16 & 14 & 3 & 12 & 32 \\  
& CONV3 & 14 & 6 & 3 & 32 & 64 \\
& FC & 1 & 1 & 1 & 2304 & 10 \\ \midrule
\bottomrule
\end{tabular}
\end{table}

\subsection{Benchmark and Baseline}
\subsubsection{Benchmarks}
In this work, we evaluate hardware performance by using two widely adopted neuromorphic vision datasets: DVS-Gesture and NMNIST. 

\textbf{DVS-Gesture:}The Dynamic Vision Sensor Gesture dataset \cite{DVS128_gesture_dataset} consists of 11 gestures from  multiple human subjects  as seen through a dynamic vision sensor, an event-based camera that responds to localized changes in brightness. Using this dataset, we train two networks, DVS-Gesture Medium and DVS-Gesture Large with different network capacities to perform the gesture classification task.

\textbf{NMNIST:} The Neuromorphic MNIST dataset(N-MNIST)\cite{NMNIST} is the neuromorphic version of the popular  MNIST dataset \cite{MNIST}. 
It consists of the same 60,000 training and 10,000 testing samples as in the original MNIST dataset.
The N-MNIST dataset was captured by mounting the ATIS sensor on a motorized pan-tilt unit and having the sensor move while it views MNIST examples on an LCD monitor. 


Based on the two datasets, we train three spiking neural network models under different settings. The configurations of these three networks are summarized in Table~\ref{table:config}. In the table, H, R, E, C, and M represent input feature map(IFmap) size, output feature map(OFmap) size, filter size, IFmap channel, and OFmap channel, respectively.

\subsubsection{Baseline}
We use the recent work of parallel time batching\cite{PTB} as the baseline to evaluate hardware performance in terms of latency and energy. The baseline employs time-domain parallel processing to perform synaptic integration under a similar memory hierarchy and systolic array setting. The approach utilizes a time-domain packing strategy without weight reuse across post-synaptic neurons and applies an optimal fixed time window size for different layers based on exhaustive search. 



\section{Results}
Because our architecture is specifically designed to take advantage of the high degree of sparsity present in SNNs, the efficiency demonstrated is highly dependent on the dataset and network used, as well as the trained sparsity of the individual layers of the network. 

\subsection{Results based on Synthetic Input Firing Rates}
\begin{figure}[ht]
    \centering
    \includegraphics[width=0.5\textwidth]{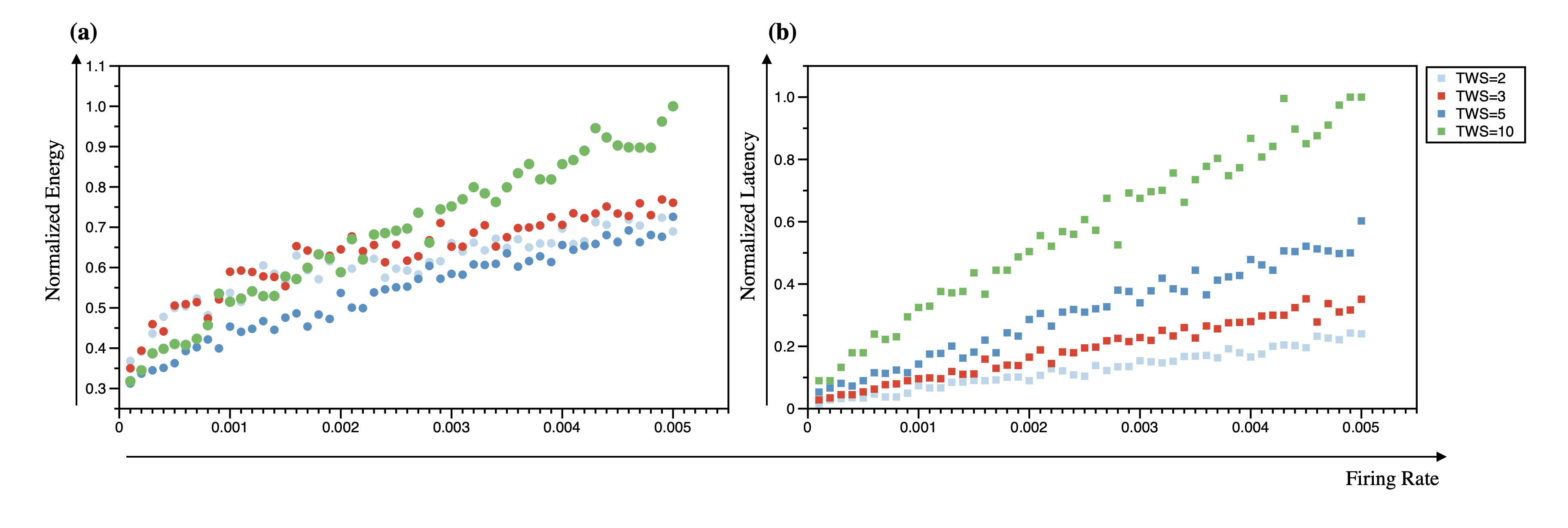}
    \caption{The normalized energy consumption and latency under different spike firing rate and TWS in CONV1 layer of N-MNIST.}
    \label{fig:Sparsity_EL}
\end{figure}

First, to gain an insight on the performance of  the proposed SpikeX architecture, we evaluate its performance in accelerating a given SNN layer for which the presynaptic input activations are generated randomly with an average firing rate in typical observed ranges.   
Figure~\ref{fig:Sparsity_EL} illustrates the impact of pre-synaptic neuron firing rates on normalized energy consumption and latency in the first CONV layer of the SNN trained on NMNIST. As the firing rate decreases from 0.5\% to 0.01\%, energy consumption decreases by 46.6\%, 54.0\%, 56.9\%, and 68.2\% when TWS equals 2,3,5,10, respectively. Meanwhile, latency is also significantly associated with firing rate, and so is reduced by 93.9\%, 92.0\%, 91.1\%, and 90.4\% respectively. Moreover, we observe the selection of an optimal TWS is not trivial, being highly dependent on the input sparsity and the parameters of each individual layer. For example, in the CONV1 layer of NMNIST, when pre-synaptic neuron firing rates are 0.5\%, a smaller TWS is preferred to remove latency introduced by larger TWS values. Conversely, when firing rates are extremely low at around 0.01\%, a larger TWS with a size of 10 is preferred to facilitate more multi-bit weight reuse. These results underscore the importance of the proposed SpikeX-HAS approach. 

\begin{figure}[htbp]
    \centering
    \includegraphics[width=0.5\textwidth]{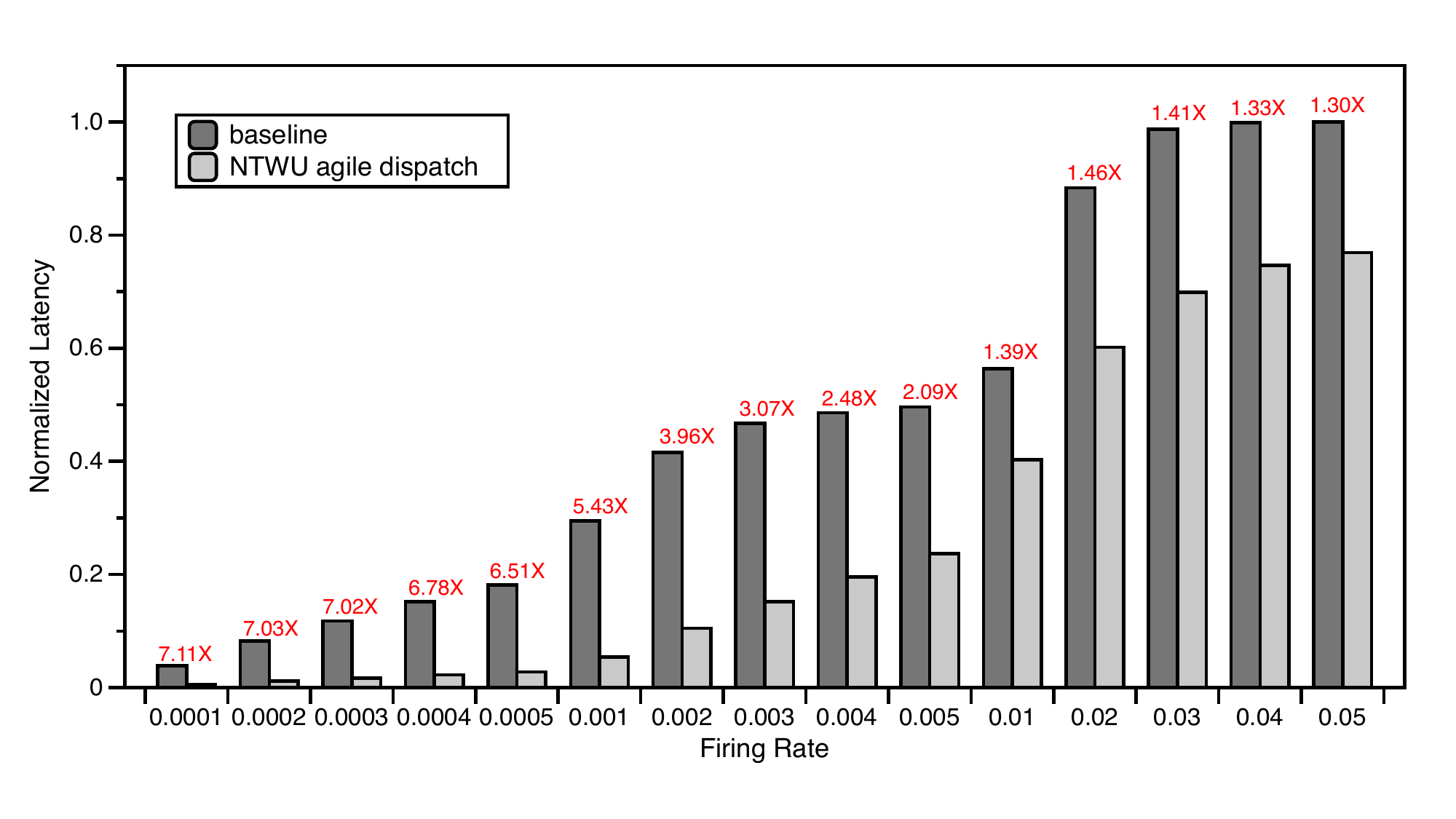}
    \caption{The normalized latency under different spike firing rate of CONV2 in DVS-Gesture Medium, with- and without agile NTWU dispatch.}
    \label{fig:Agile}
\end{figure}

Under various input firing rates, Figure~\ref{fig:Agile} shows that employing agile NTWU dispatch in the proposed SpikeX architecture significantly improves latency when compared with the basedline PTB architecture \cite{PTB}.  As the firing rate decreases, the advancement of agile NTWU dispatch is enhanced. Because as the overall firing rate decreases, the firing rate of a fixed neuron or a fixed group of neurons is also decreasing, so that PTB baseline cannot fully utilize systolic array in the traditional work\cite{PTB}. When the firing rate is 0.01\%, there is a significant latency improvement at 7.11$\times$ compared with baseline. When the firing rate is high at 5\%, the temporal NTWUs become more dominant, agile NTWU dispatch also has a 1.3X latency improvement compared with the baseline because it can still fully utilize PE array to efficiently handle temporal parallel work and spatially parallel work simultaneously. Furthermore, we observe that agile NTWU dispatch in SpikeX strong reduction in latency as the firing rate decreases.

\subsection{Impacts on Time Window Size}
\begin{figure}[ht]
    \centering
    \includegraphics[width=0.5\textwidth]{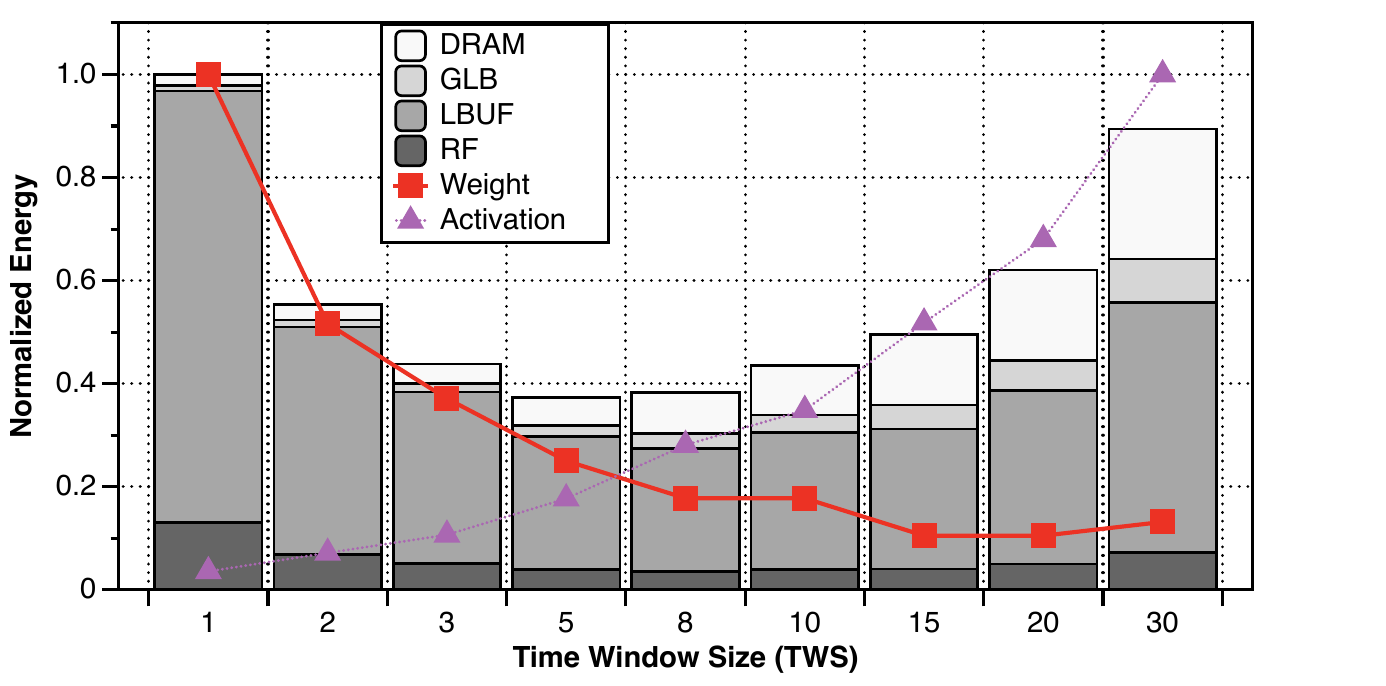}
    \caption{The impact of time window size on normalized energy in CONV2 of DVS-Gesture Large network.}
    \label{fig:TWS}
\end{figure}

As can be seen in Figure~\ref{fig:TWS}, under the actual inputs to a known network layer in the SNN model, the energy is highly dependent on the Time Window Size.  Similarly as in Figure~\ref{fig:Sparsity_EL}, latency is highly dependent on TWS.  

\begin{figure}[ht]
    \centering
    \includegraphics[width=0.4\textwidth]{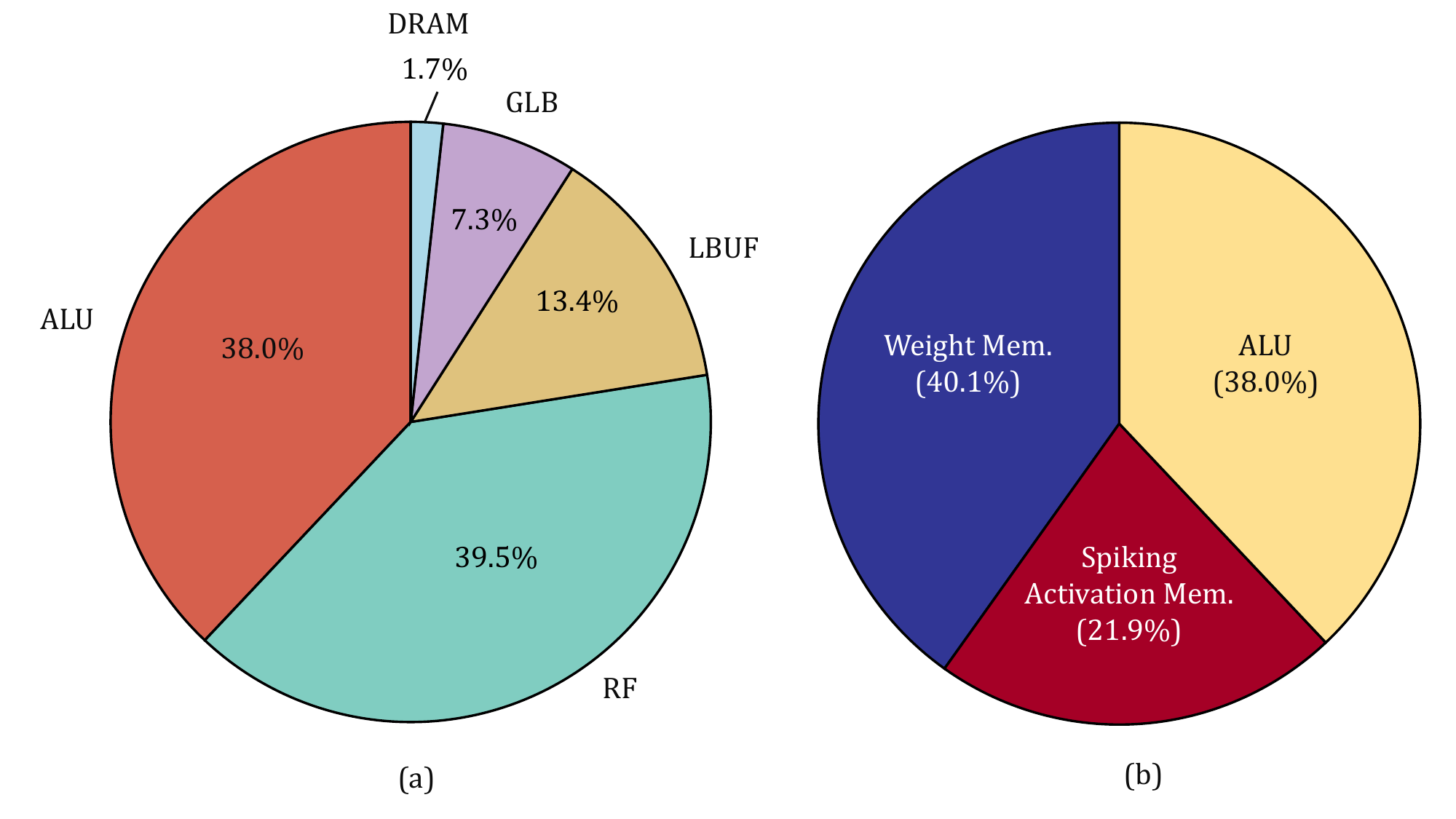}
    \caption{\textcolor{black}{Energy consumption breakdown on FC layers when TWS=4 (a) by components—ALU, RF, LBUF, GLB, and DRAM. (b) by usage—weight memory access, spiking activation memory access, and ALU computation.}}
    \label{fig:energy_breakdown}
\end{figure}

\textcolor{black}{
Figure~\ref{fig:energy_breakdown}(a) and (b) illustrate the energy consumption breakdown of SpikeX when executing fully-connected layers. By component, memory accounts for 62\% of the total energy consumption, with a detailed distribution as follows: RF (39.5\%), LBUF (13.4\%), GLB (7.3\%), and DRAM (1.7\%). The ALU contributes the remaining 38\%. By usage, 40.1\% of the energy is consumed by weight memory access, while 21.9\% is consumed by spiking activation memory access. When the time window size is increased from 4 to 8, the energy consumption for weight memory access decreases to 36.19\%, whereas the energy consumption for spiking activation memory access increases to 30.65\%.
}

\subsection{Comprehensive Evaluation}

The network-hardware co-optimization framework automatically determines the key hardware architectural parameter, TW size, for each SNN layer, during training with a chosen hardware/accuracy tradeoff hyperparameter, $\beta$. We comprehensively evaluate the effects of the proposed techniques and their aggregated improvements while comparing to the PTB baseline \cite{PTB}, for which
the time window size  is chosen to the optimal one based on exhaustive search in a wide range. 
The systolic array size is set to 8x8, a near-optimal value.  

\begin{figure*}[ht]
    \centering
    \includegraphics[width=1\textwidth]{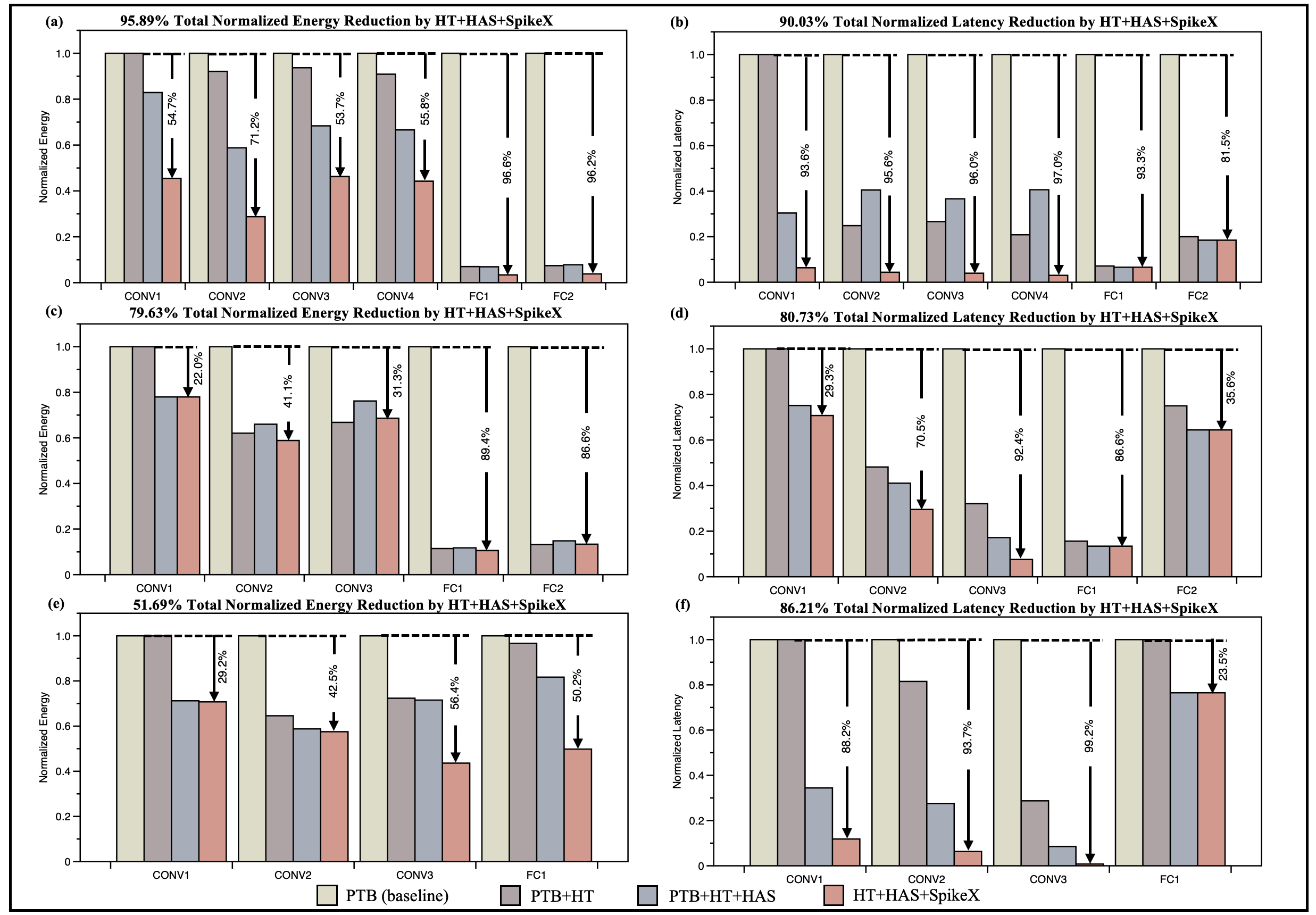}
    \caption{The normalized latency and energy with- and without SpikeX, SpikeX-HT and SpikeX-HAS, in each dataset. \textbf{(a),(b): DVS-Gesture-Medium, (c),(d): DVS-Gesture-Large and (e)(f): NMNIST.} SpikeX with HT and HAS improves total energy dissipation and latency by DVS-Gesture-Medium: 24.38 $\times$ and 10.29$\times$, DVS-Gesture-Large: 4.91$\times$ and 5.19$\times$, and NMNIST: 2.07$\times$ and 7.25$\times$, over the baseline. The performance of CONV1 is kept the same after applying PTB+HT due to predefined inputs in the validation set.}
    \label{fig:Eval_fig3}
\end{figure*}

Figure~\ref{fig:Eval_fig3} compares the energy and latency with and without hardware-aware training: SpikeX-HT, hardware architecture search: SpikeX-HAS, and SpikeX architecture. Note that given the generality of our optimization techniques, they can be applied to the PTB architecture as well. 

\subsubsection{Performance of SpikeX-HT and SpikeX-HAS}
\textbf{Energy Dissipation-} Hardware training (Spike-HT or HT) significantly reduces energy dissipation by 11.13$\times$, 4.89$\times$ and 1.21$\times$ on three different SNNs on DVS-Medium, DVSG-Large and NMNIST. HT reduces energy consumption except for the input layer where the spatial and temporal sparsity cannot be controlled and is obtained from the external environment. However,  the spiking activities in the hidden and output layers are effectively optimized, for which there is a maximum energy reduction of 14.25$\times$ in a layer. 

When hardware architectural optimization (SpikeX-HAS or HAS) is applied, the optimal TWS for each layer is automatically determined during training. This results in a further reduction of total energy dissipation, with an 8.5\% and 9.8\% improvement on DVS-Gesture Medium and NMNIST. For some layers, the energy consumption may slightly increase after optimizing HW architectural parameters, due to the trade-off between energy and latency. Under these cases, sacrificing energy efficiency slightly can lead to better latency by switching to another TWS. Specifically, for the DVS-Gesture Large model,  HAS increases energy  by 10.5\%, but leads to a 14.2\% reduction of latency. The energy-latency tradeoff can be further altered by choosing a different weighting parameter between the two in the loss function. 

\textbf{Latency}- SpikeX-HT demonstrates a very significant latency reduction as compared to the baseline in all three networks, as shown in Figure~\ref{fig:Eval_fig3}. Spike-HT (or HT) offers an improvement of 9.44$\times$, 4.29$\times$, and 1.85$\times$ on the total latency of DVS Gesture Medium, DVS Gesture Large, and NMNIST. After adapting HAS, a 10.18$\times$, 5.01$\times$, and 4.17$\times$ improvement in total latency is achieved compared with the baseline. As spiking activities become sparser, less data is loaded from the memories and fewer computations are required to be performed, resulting in an improved  system with a less latency. 

\subsubsection{Performance of SpikeX with co-optimized network}
As shown in \ref{fig:Eval_fig3}, with the obtained optimal TWS from HAS, SpikeX architecture with weight tailoring and agile spatiotemporal dispatch techniques can boost energy efficiency significantly by 24.38$\times$, 4.91$\times$, and 2.07$\times$ on three distinct networks. The latency of the total network is also reduced by 10.29$\times$, 5.19$\times$ and 7.25$\times$ compared with the baseline.

\textbf{EDP Evaluation-} We use energy-delay product(EDP) to simultaneously consider the latency and energy efficiency of the overall system. We multiply the total energy dissipation and the total amount of execution time to calculate the value of EDP. 
Although \cite{PTB} demonstrates a 248$\times$ EDP improvement on average over another recent work of \cite{Lee_ICCD_2020}, this present work can have a further 15.1$\times$ to 150.87$\times$ improvement on \cite{PTB} with our proposed agile spatiotemporal dispatch and activation-induced weight tailoring as part of the SpikeX hardware architecture, and our proposed network and accelerator co-optimization techniques. 


\section{Conclusions}
In this paper, we propose SpikeX, a novel hardware architecture for sparse spiking neural network processing with weight tailoring and agile spatiotemporal dispatch techniques to improve system throughput and energy efficiency. SpikeX takes advantage of the inherent spatiotemporal unstructured sparsity in SNNs, and fully exploits multi-bit weight reuse within- and across- time windows and post-synaptic neurons. To optimize deployment efficiency on SpikeX, we propose a novel hardware-aware training methodology, SpikeX-HT, to effectively improve energy efficiency and throughput introduced by firing sparsity from spiking activities. Additionally, we propose a joint network-hardware architecture co-optimization(Spike-HAS) scheme to optimize hardware architectural parameters by searching the optimal time window size(TWS) during training. Using this, we demonstrate a set of tradeoffs between accuracy and EDP, with at least one simultaneous increase in accuracy and decrease in EDP for each network and dataset tested. Our efficient SpikeX hardware architecture and end-to-end network/accelerator co-design approach offer a significant reduction of 15.1$\times$$-$150.87$\times$ in energy-delay-product without sacrificing accuracy. Altogether, our work demonstrates an efficient architecture for the deployment of Spiking Neural Networks, alongside an SNN training methodology which optimizes the network for such a hardware accelerator. \\

\section{Acknowledgement}
This material is based upon work supported by the National Science Foundation under Grants No. 1948201 and No. 2310170.

\bibliographystyle{IEEEtranS}
\bibliography{refs}

\end{document}